\documentclass[conference,10pt]{IEEEtran}
\usepackage[T1]{fontenc}
\usepackage[utf8]{inputenc}
\usepackage[expansion=false]{microtype}
\usepackage{cite}
\usepackage{amsmath,amssymb}
\usepackage{graphicx}
\usepackage{helvet}
\usepackage{xcolor}
\usepackage{booktabs}
\usepackage{array}
\usepackage{listings}
\lstdefinestyle{rtlcase}{
  language=Verilog,
  basicstyle=\ttfamily\small,
  commentstyle=\color{black!72},
  columns=fullflexible,
  keepspaces=true,
  showstringspaces=false,
  aboveskip=3pt,
  belowskip=3pt
}
\usepackage[hidelinks]{hyperref}

\newif\ifanonymous
\newcommand{\gym}{SiliconGym}
\newcommand{\papertitle}{SiliconGym: A Versatile Reinforcement Learning Environment for Agentic EDA}
\hypersetup{
  pdftitle={\papertitle},
  pdfauthor={},
  pdfsubject={Reinforcement learning environments for hardware design and verification},
  pdfkeywords={electronic design automation, reinforcement learning, hardware verification}
}

\IEEEoverridecommandlockouts
\hypersetup{pdfauthor={Hejia Zhang, Sheng Lu, Yitong Ma, Zhongming Yu, Zhijing Wu, Wentao Ni, Chia-Tung Ho, Jishen Zhao}}
\begin{document}
\title{\papertitle}
\author{
\IEEEauthorblockN{
Hejia Zhang\textsuperscript{1,*},
Sheng Lu\textsuperscript{2,*},
Yitong Ma\textsuperscript{1},
Zhongming Yu\textsuperscript{1},\\
Zhijing Wu\textsuperscript{1},
Wentao Ni\textsuperscript{1},
Chia-Tung Ho\textsuperscript{3},
Jishen Zhao\textsuperscript{1}\thanks{\textsuperscript{*}Equal contribution.}}
\IEEEauthorblockA{
\textsuperscript{1}University of California San Diego, La Jolla, CA, USA\\
\textsuperscript{2}Georgia Institute of Technology, Atlanta, GA, USA\\
\textsuperscript{3}NVIDIA, Santa Clara, CA, USA\\
\textsuperscript{1}\texttt{\{hez024,yim030,zhy025,zhw163,w2ni,jzhao\}@ucsd.edu}\\
\textsuperscript{2}\texttt{slu375@gatech.edu}\quad
\textsuperscript{3}\texttt{chiatungh@nvidia.com}}
}

\maketitle
\begin{abstract}
Reinforcement learning (RL) with electronic design automation (EDA) feedback is advancing large language models for hardware design and verification.
Extending these gains to hardware-engineering agents requires environments that span tasks, preserve state across tool calls, and support large hierarchies alongside parallel rollouts.
We present SiliconGym, the first versatile, cross-task and cross-toolchain RL environment for hardware design and verification.
By decoupling versioned workspaces, budgeted execution, and acceptance-gated scoring, SiliconGym scales from module-level RTL, testbench, and PPA tasks to multi-round repository repair and full-hierarchy SoC flows.
Streaming dispatch overlaps tool checks with generation, finishing a 256-answer RL batch 21.9\% faster than native ChipSeek.
Using interchangeable open-source and commercial backends, we show that open-source-trained functional correctness transfers cleanly to commercial simulation, while paired netlist evaluation reveals when open-source PPA gains survive commercial synthesis.
\end{abstract}

\bstctlcite{IEEEexample:BSTcontrol}
\section{Introduction}
\label{sec:introduction}

\begin{figure}[!t]
  \centering
  \includegraphics[width=\columnwidth]{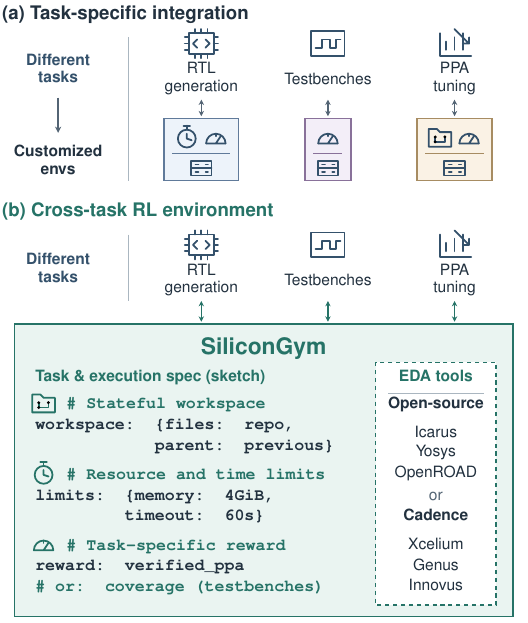}
  \caption{Task-specific harnesses with uneven state and resource controls (a)
  versus \gym{}'s shared cross-toolchain environment (b). Folder, clock, and gauge
  icons denote workspace state, limits, and rewards. Arrows carry candidates and
  feedback/rewards between trainers/agents and environments. The sketch shows
  example workspace inheritance, budgets, and acceptance-gated rewards.}
  \label{fig:teaser}
\end{figure}

Large language models (LLMs) have improved register-transfer-level (RTL) generation on benchmarks such as VerilogEval and RTLLM~\cite{verilogeval,verilogeval2,rtllm}.
Reinforcement learning (RL) drives recent gains in RTL generation, verification, and power, performance, and area (PPA) optimization using executable electronic design automation (EDA) feedback~\cite{codevr1,provr1,chipseek} or learned rewards derived from simulation~\cite{verirl_teng}.
Concurrent work also trains models to invoke EDA tools and revise module-level RTL~\cite{rtlinteract}.
The next step toward practical hardware engineering extends beyond standalone modules: agents must inspect tool feedback, revise multiple files, and test changes within existing design hierarchies~\cite{hwebench,horizon}.
Training such agents requires environments that connect repeated tool calls and parallel rollouts to workflows ranging from module simulation to full-system implementation.

Existing task-specific EDA-RL pipelines tie workspace handling, tool invocation, and reward scripts to individual tasks and toolchains, leading to \textbf{repeated integration effort} when workflows change (Fig.~\ref{fig:teaser}(a)).
As rollout concurrency and optimization objectives expand, \textbf{fragile reward evaluation} corrupts the learning signal: resource interference changes scores, and incomplete checks reward invalid designs.
In a controlled stress replay of a simulation-based verifier~\cite{codevr1}, a 23-line generated RTL candidate produced a 16.4\,GiB whole-container memory peak, and four identical copies scored $[1,0,1,1]$ in a shared-memory batch with out-of-memory kills.
Simulation acceptance alone overstates PPA gains: 31 of 54 passing candidates with apparent area reductions above 5\% failed formal, interface, or initialization checks.
Beyond single-toolchain reliability, a shared cross-backend environment makes it possible to test a key question for researchers limited to open-source (OSS) tools: whether OSS-trained capabilities~\cite{codevr1,chipseek} carry over to commercial EDA evaluation.

Building such an environment requires resolving three technical challenges.
\textbf{Stateful interaction across design scales.}
Single-shot submissions and iterative repository edits require different workspace lifetimes.
The environment must preserve editable files and public diagnostics across revisions while protecting private tests and carrying project dependencies from individual modules to full design hierarchies.
\textbf{Bounded execution and valid rewards.}
Multi-stage EDA jobs have artifact dependencies and wide variation in runtime and memory use.
Parallel rollouts need explicit resource budgets and failure outcomes, while task-specific acceptance checks must verify functional and implementation contracts without adding prohibitive overhead.
\textbf{Toolchain-independent task semantics.}
When task logic hardcodes tool invocations and report formats, changing a backend risks changing the task or reward definition as well.
Studying transfer from open-source to commercial feedback therefore requires interchangeable backends and paired scoring under explicit task rules.

Addressing these challenges, we present \gym{}, the first versatile, cross-task and cross-toolchain RL environment for hardware design and verification, with three contributions:
\begin{itemize}
  \item \textbf{Versatile, stateful interaction across design scales.}
  Versioned workspaces and private-oracle isolation let one protocol span five repository repairs, multi-round feedback revisions, and full-hierarchy Earlgrey and OpenTPU reference flows through global placement on OSS and commercial backends.
  \item \textbf{Budgeted parallel execution and reward validation.}
  Explicit budgets preserve standalone-verifier scores in six matched stress replays, while qualification and acceptance gates reject invalid PPA gains.
  On a 256-answer ChipSeek replay, SiliconGym returns 245 rewards before generation finishes, reducing batch time by 21.9\% relative to the native harness.
  \item \textbf{RL transfer from open-source to commercial tools.}
  With interchangeable backends, we find that functional correctness transfers cleanly across simulators on VerilogEval ($<0.2$ percentage-point aggregate difference).
  OSS PPA training yields modest commercial-scored gains; netlist analysis explains why commercial synthesis absorbs other OSS-rewarded RTL rewrites.
\end{itemize}

\begin{figure*}[!t]
  \centering
  \includegraphics[width=\textwidth]{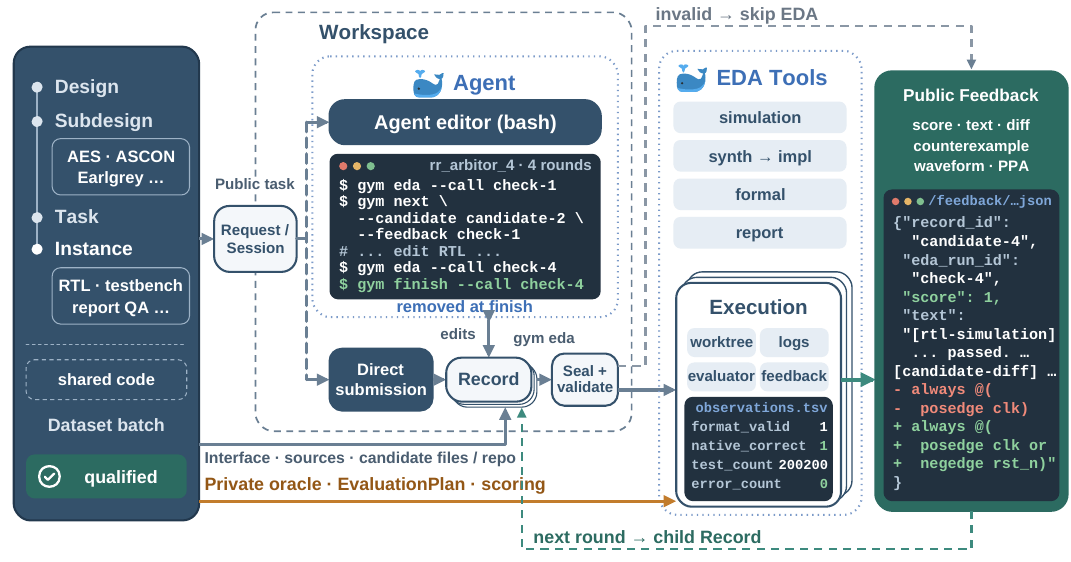}
  \caption{\gym{} architecture. Qualified task definitions feed direct submissions or agent-edited Records. Submitted candidates are sealed and validated before EDA tools perform simulation, synthesis, formal checks, or report analysis. Public feedback supports subsequent rounds; command and output excerpts illustrate the interaction.}
  \label{fig:architecture}
\end{figure*}

\section{Background and Related Work}
\label{sec:related-work}

\subsection{Learning and Agentic Search with EDA Feedback}
EDA feedback guides both training and inference: for RTL generation, VeriRL trains a reward model from simulation-derived preferences to guide Reinforce++ updates~\cite{verirl_teng}, while CodeV-R1 and ChipSeek obtain simulation and PPA rewards from tool execution~\cite{codevr1,chipseek}.
The concurrent RTL-Interact preprint trains tool-using policies for module-level RTL tasks and identifies repository-level training environments as future work~\cite{rtlinteract}.
Beyond RTL design, LLM4Cov learns testbench generation from execution-validated offline trajectories~\cite{llm4cov}, and CHORUS trains complementary coverage experts with dense-reward RL before merging or distilling them~\cite{chorus}.
At inference time, feedback drives testbench correction through functional self-validation in CorrectBench~\cite{correctbench} and joint RTL/synthesis search with isolated execution and independent re-evaluation in RTLScout~\cite{rtlscout}.
HORIZON extends the search setting to repository-level hardware evolution using executable acceptance conditions and isolated Git worktrees~\cite{horizon}.
\gym{} supplies a common candidate--feedback runtime for these trainers and search agents across tasks and backends.

\subsection{Hardware Benchmarks and Evaluation Infrastructure}
VerilogEval checks functionality through simulation~\cite{verilogeval,verilogeval2}, RTLLM includes design quality~\cite{rtllm}, and VeriThoughts uses formal verification~\cite{verithoughts}. Other benchmarks target financial FPGA latency~\cite{finhardbench} and RTL debugging~\cite{chipbench}.
Broader workflows require additional evaluation contexts: CVDP covers agentic and non-agentic design/verification tasks~\cite{cvdp}, HWE-Bench uses project-native regressions for repository repair~\cite{hwebench}, and EDAFlowQA tests reasoning over tool logs~\cite{edaflowqa}.
TuRTLe consolidates how generated RTL is scored through a common correctness, synthesis, and PPA evaluation flow~\cite{turtle}.
\gym{} makes benchmark-defined checks usable within budgeted parallel rollouts and stateful agent interaction.

\subsection{Reusable Environments and RL Systems}
Common action--observation interfaces decouple learning algorithms from environment implementations, as in Gymnasium~\cite{gymnasium}; CompilerGym~\cite{compilergym} and ArchGym~\cite{archgym} adapt this pattern to compiler optimization and architecture simulation.
For language models, HybridFlow distributes RL computation~\cite{hybridflow}, ChipMATE trains cooperating RTL and reference-model agents~\cite{chipmate}, and OpenEnv provides isolated synchronous and asynchronous interaction~\cite{openenv}.
An OpenEnv-based RTL environment supports multi-step, multi-file interaction and concurrent sessions for accelerator primitives using Icarus, Yosys, and optional SymbiYosys~\cite{verirl_env}.
\gym{} extends this RTL-focused setting to testbench and report tasks, with interchangeable commercial backends and configuration qualification that checks acceptance and isolation independently of the learning algorithm.

\section{SiliconGym Architecture}
\label{sec:architecture}

SiliconGym separates candidate state, task and reward specifications, and budgeted execution so that interaction mode, design scale, and tool choice can vary independently (Fig.~\ref{fig:architecture}).

\subsection{Versioned Workspaces and Oracle-Isolated Interaction}
Parallel trainers submit independent candidates through \emph{Direct submission}; iterative agents revise workspaces through \emph{Agent editor (bash)}.
Both paths use a \emph{Record} to bind each reward to one candidate version despite their different workspace lifetimes.
The task instance declares editable paths and readable public context: a generation task may accept one RTL file, whereas a repair task permits edits, additions, and deletions across a repository.

An agent can edit and self-test while its Record is open.
Submission invokes \emph{Seal + validate}: the system freezes and hashes the candidate, checks its interface, and evaluates that snapshot.
A revision creates a child Record that inherits permitted files and links to the parent's public feedback, preserving the relationship between edits, diagnostics, and rewards across rounds.
The agent can inspect task-selected waveforms, coverage reports, or counterexamples, while hidden reference tests and scoring bindings remain in the tool-side \emph{Private oracle}.

\subsection{Toolchain-Decoupled Task and Reward Specifications}
The evaluation pipeline separates what a task rewards from which tool executes it.
For an RTL-optimization candidate replacing a module in a fixed design snapshot, \emph{instance binding} maps the edited file to the flow's expected location, while an \texttt{EvaluationPlan} selects simulation, equivalence, and synthesis configurations.
An \emph{execution adapter} stages the inputs and invokes the tools under their budgets; an \emph{evidence collector} translates tool-specific reports into structured observations such as mismatch counts, proof depth, and cell area (\texttt{observations.tsv} in Fig.~\ref{fig:architecture}).
The \emph{reward evaluator} applies the task's interface, initialization, and equivalence gates before converting an accepted area reduction into reward.

Changing toolchain therefore changes the execution configuration and report collector while preserving the candidate interface and task-level reward rule.
Artifact dependencies govern multi-stage execution: placement consumes an accepted synthesis result, while independent simulation and synthesis may run concurrently.
Proof depth and status let reward rules distinguish a bounded sequential pass from a complete proof.

\begin{table*}[!t]
  \centering
  \caption{Cross-task and cross-scale coverage. The catalog counts definitions (defs.). Execution counts cover the training runs, cases, and reference flows reported here.}
  \label{tab:coverage}
  \footnotesize
  \setlength{\tabcolsep}{3pt}
  \renewcommand{\arraystretch}{1.12}
\begin{tabular}{@{}>{\raggedright\arraybackslash}p{.07\textwidth}>{\raggedright\arraybackslash}p{.20\textwidth}>{\raggedright\arraybackslash}p{.19\textwidth}>{\raggedright\arraybackslash}p{.17\textwidth}>{\raggedright\arraybackslash}p{.09\textwidth}>{\raggedright\arraybackslash}p{.22\textwidth}@{}}
  \toprule
  \textbf{Category} & \textbf{Task / workload} & \textbf{Scale / interaction} & \textbf{Evaluation stages} & \textbf{Backends} & \textbf{Catalog / execution scope} \\
  \midrule
  \textbf{Training} & RTL generation / CodeV & Module / parallel RL & Simulation & OSS, Comm. & 3,133 defs.; $2\!\times\!200$ updates \\
  & PPA optimization / ChipSeek & Module / parallel RL & Sim., synthesis & OSS, Comm. & $2\!\times\!200$ updates; 32 slots \\
  \midrule
  \textbf{Cases} & Repository repair / HWE & Repository / Bash agent & Build, regression & OSS & 5 repairs; 1 Record each \\
  & Feedback RTL repair & Module / multi-round agent & Sim., waveforms & OSS & Arbiter + 3 WaveDebug repairs \\
  & Formal RTL / VeriThoughts & Module / controls, repair & Bounded equivalence & OSS & 5 qualified; 1 two-round repair \\
  \addlinespace[1.5pt]
  & Testbench / CorrectBench & Module / TB controls & Sim. (golden/mutant) & OSS & 5 qualified cases \\
  & Coverage testbench / CVDP & Module / trainer integration & Sim., coverage & Comm. & 4 qualified; 1 RL update \\
  \addlinespace[1.5pt]
  & RTL optimization / RTLScout & Module / tool, agent trials & Synthesis, placement & OSS & Tool trials; 1-round agent trial \\
  & PPA validity / CodeV-PPA & Module / candidate replay & Sim., formal, synthesis & OSS & 289 defs.; 508-candidate replay \\
  \addlinespace[1.5pt]
  & Report QA / EDAFlowQA & Reports / answer controls & Parsing, scoring & --- & 5 qualified; cached logs \\
  \midrule
  \textbf{Flows} & Earlgrey / OpenTPU & SoC, accelerator / reference & Sim., synth., global place & OSS, Comm. & Both designs, both backends \\
  \bottomrule
\end{tabular}
\end{table*}

\subsection{Budgeted Execution and Configuration Qualification}
Configurations budget EDA execution independently of the agent's editing session, limiting the resources a candidate can consume during parallel rollouts.
The reference container adapter uses container memory and process controls; subprocess adapters apply supported process limits and periodically check log and workspace usage.
Both supervise termination and retain candidate, dependency, infrastructure, and cancellation outcomes alongside the score.
Trainers can distinguish budget exhaustion, a failed test, and an unavailable tool.

Budgets alone cannot ensure that a task's checks or visibility rules are correct.
Configuration qualification exercises five facets for an exact definition and configuration: executability, reference acceptance, negative controls, oracle isolation, and agent visibility.
The \emph{qualified} badge in Fig.~\ref{fig:architecture} applies to both the tested task and tool definitions. A change to either triggers requalification.

\section{System Evaluation}
\label{sec:evaluation}

We evaluate versatility, reward fidelity, and runtime cost before studying cross-toolchain RL transfer in Section~\ref{sec:toolchains}.

\subsection{RQ1: Task Versatility and Design Scale}
Table~\ref{tab:coverage} separates model training, capability demonstrations, and reference flows.
The catalog contains 3,470 definitions; 3,133 CodeV instances pass four qualification facets, and 24 across task families pass all five, including negative controls.

\paragraph{Training and verification}
Binary testbench checks, continuous coverage and PPA rewards, and report QA share one candidate--feedback schema (Table~\ref{tab:coverage}).
Task evaluators set acceptance criteria and scalar rewards, so trainers can change objectives through the same interface.

\paragraph{Stateful repair}
Five model-authored repairs pass targeted regressions in complete Ibex, CVA6, Caliptra, Rocket Chip, and XiangShan repository workspaces (Table~\ref{tab:capabilities}).
Each run uses multiple Bash actions within one Record; Ibex edits five RTL and three documentation files and performs five self-tests before submission.
Directory-binding controls additionally exercise file additions and deletions.

Parent--child Records extend this interaction across scored revisions.
In the \texttt{arbiter} case, the model repairs its own failed candidate using public feedback, improving its score from zero to one across two evaluations (with CLI commands relayed).
Three WaveDebug cases and the VeriThoughts \texttt{pagesel} case repair seeded failures using inherited waveforms or a nine-step formal counterexample.
Agents can edit and self-test within a turn. Sealing binds the reward to that snapshot and carries the diagnostics into the next turn.

\begin{table}[!t]
  \centering
  \caption{Five single-Record repairs, each passing the named regression. Edited files count source/document changes. SV: SystemVerilog; RoT: root of trust.}
  \label{tab:capabilities}
  \footnotesize
  \setlength{\tabcolsep}{2pt}
  \renewcommand{\arraystretch}{1.12}
  \begin{tabular}{@{}>{\raggedright\arraybackslash}p{.18\columnwidth}>{\raggedright\arraybackslash}p{.23\columnwidth}>{\centering\arraybackslash}p{.11\columnwidth}>{\centering\arraybackslash}p{.11\columnwidth}>{\raggedright\arraybackslash}p{.30\columnwidth}@{}}
  \toprule
  \textbf{Repository} & \textbf{Lang. / domain} & \textbf{\shortstack{Edited\\files}} & \textbf{\shortstack{Bash\\cmds}} & \textbf{Targeted regression} \\
  \midrule
  Ibex & SV / RISC-V & 8 & 17 & Debug-mode PMP access \\
  \addlinespace[1.5pt]
  CVA6 & SV / RISC-V & 1 & 8 & Compressed decode legality \\
  \addlinespace[1.5pt]
  Caliptra & SV / RoT & 1 & 7 & Key-vault error retention \\
  \addlinespace[1.5pt]
  Rocket Chip & Chisel / RISC-V & 1 & 16 & JTAG asynchronous reset \\
  \addlinespace[1.5pt]
  XiangShan & Chisel / RISC-V & 1 & 13 & Virtual interrupt selection \\
  \bottomrule
\end{tabular}

\end{table}

\paragraph{Full-hierarchy execution}
Reference flows for Earlgrey and OpenTPU reach global placement on open-source and commercial backends, using ASAP7 and approximate memory views under backend-specific resource configurations.
Earlgrey retains 19 memory macros and passes two firmware tests; its open and commercial input-to-feedback-and-cleanup times are 13,216 and 14,048\,s, respectively.
OpenTPU contains 212,256 generic logic/register cells and 34 memory cells totaling 34.5\,Mbit.
Five workloads pass on both chains, and a Genus-netlist zero-delay check passes 1,656 cycles.
A candidate can thus reach global placement with its enclosing hierarchy and memory configuration intact.

\subsection{RQ2: Resource Budgets and Reward Fidelity}
\label{sec:validity}
Timeouts alone do not protect concurrent rollouts from memory pressure.
In targeted stress probes, the baseline verifier replay~\cite{codevr1} uses a 30-second watchdog and process-tree termination, yet shared-memory pressure changes scores (Section~\ref{sec:introduction}).
In a shared-container probe, a 40\,GiB trainer surrogate with elevated OOM-kill priority exits 137 alongside six cgroup OOM kills.
Three isolated timeout probes each retain a defunct process and work directory after two seconds.

Under matched budgets---a 1\,GiB container and 768\,MiB tool address-space limit---all six Gym/native replay pairs agree on their scores.
The memory-heavy candidate passes with sufficient memory but scores zero under tight limits.
The Gym preserves the standalone verifier's reward at both operating points, including when the tool runs out of memory.

Across 508 comparable simulation-passing candidates with valid PPA from the CodeV-PPA corpus, 54 appear to reduce area by over 5\%.
Acceptance gates reject 31 of these 54 candidates (57.4\%): 24 yield formal counterexamples, five violate the ASIC initialization contract, and two mismatch port interfaces.
Of the remaining candidates, 14 pass 20-step bounded equivalence, eight have complete combinational proofs, and one reaches the formal timeout.
Retaining proof scope lets a trainer require stronger checks for sequential designs whose errors may emerge beyond a chosen bound, instead of treating every bounded pass as complete equivalence.
Two simulation-passing candidates illustrate how these apparent gains arise.
For \texttt{audio} (CodeV-68851), a rewrite replaces the reference's 33-bit counters and bit-selected output with narrower counters, equality tests, and control flags.
Mapped area falls by 34.9\%, but the rewrite also changes the counter and reset behavior: a formal trace finds an output mismatch at step 19.
For \texttt{sixtycounter} (CodeV-60023), a candidate narrows a 28-bit counter to 21 bits and adds an \texttt{initial} assignment; its measured area falls by 10.9\%.
The initialized state is outside the accepted ASIC implementation contract, so this candidate fails the initialization check.

\subsection{RQ3: Runtime Cost and Parallel Efficiency}
\label{sec:replay}
We compare SiliconGym with ChipSeek's native harness on the first 256-answer training batch using Icarus and Yosys/OpenSTA.
All three modes replay the same generation schedule with 32 warm workers and the same per-task tool limits.
Native waits for the last answer before starting EDA. Gym dispatches each answer as it arrives. To measure what early dispatch contributes, we also run Gym with batch submission.

\begin{figure}[!t]
  \centering
  \includegraphics[width=\columnwidth]{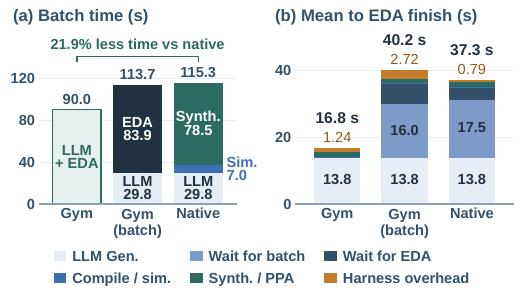}
  \caption{ChipSeek batch~1. (a) Gym overlaps generation and EDA. The two stacks show successive batch phases. Native's synthesis segment extends through cleanup. (b) Mean time to EDA completion over 256 answers. Bold totals appear above each stack. The top segment gives harness time for dispatch, setup, and worker bookkeeping. \emph{LLM Gen.} includes waiting for generation to start. Tool stages include container invocation.}
  \label{fig:timing}
\end{figure}

\paragraph{Where Gym gains time}
Gym finishes in 90.0\,s versus 115.3\,s for native, a 21.9\% reduction in generation-plus-reward time (Fig.~\ref{fig:timing}a).
Forced-batch Gym takes 113.7\,s, close to the native replay recorded under higher host load. Switching only Gym's submission mode saves 23.7\,s with identical rewards.
By the time generation ends at 29.8\,s, Gym has returned 245 rewards. Native returns all scores together after the last synthesis finishes.
Scores match on 255 of 256 answers. On the exception, native's float-only check drops an integer-valued PPA bonus ($+3$) that Gym retains.

\paragraph{Where Gym pays for managed execution}
Dispatch, setup, and worker bookkeeping average 1.24\,s per answer in streaming Gym, 2.72\,s in forced-batch Gym, and 0.79\,s in native (Fig.~\ref{fig:timing}b).
Mean compilation, simulation, and synthesis time totals 1.6--1.8\,s per answer across the three modes. Only 106 answers reach synthesis. Each of the other 150 contributes zero synthesis time to the mean, even though native holds its score until the whole batch finishes.
Releasing all answers together makes a Gym answer wait 6.09\,s on average for bridge pickup and a worker slot. Streaming cuts that wait to 0.17\,s and lets evaluation proceed while the LLM generates the remaining answers.

\paragraph{Synthesis timeouts and reward coverage}
Unlike module simulation, synthesis workloads exhibit heavy-tailed runtime distributions where a few slow RTL candidates dominate batch completion.
To evaluate how \gym{} manages the trade-off between batch throughput and reward coverage under skewed EDA workload patterns, we compare the per-task limits above against tight uniform caps (5 s for compilation/simulation and 10 s for synthesis) in streaming mode
Under uniform caps, nine answers time out in synthesis and earn only the functional reward, whereas per-task limits let eight of the nine finish in 18–34 s and earn a PPA bonus.
However, the ninth runs to its 75 s cap, extending the post-generation wait from 1.5 to 60.2 s. 
Longer budgets thus recover PPA feedback on slow designs, while tight uniform caps—adopted for our PPA training runs in Section V-B—bound tail latency from unrewarded stragglers.

\section{Sensitivity to EDA Toolchains}
\label{sec:toolchains}

Can gains learned with open-source tools carry over to commercial evaluation? We change the scoring backend while holding the model answers and task rules fixed.

\begin{figure}[!t]
  \centering
  \includegraphics[width=\columnwidth]{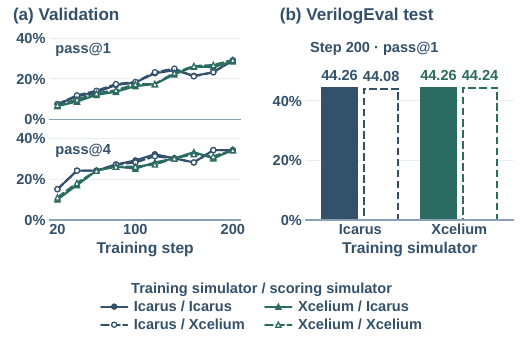}
  \caption{Both simulators score the same answers from each model. (a) Validation uses four answers per task on 99 tasks. Pass@1 is the mean sample pass rate and pass@4 is the fraction of tasks with a passing answer. (b) VerilogEval test pass@1 at step 200. Bars are grouped by training simulator. Filled marks and bars denote Icarus scoring, open marks and bars Xcelium.}
  \label{fig:functional-curves}
\end{figure}

\subsection{Functional Correctness Across Simulators}
Figure~\ref{fig:functional-curves} follows one Icarus-trained and one Xcelium-trained model from validation to the VerilogEval test.
On validation, switching the scorer gives nearly overlapping pass@1 and pass@4 curves for each model. At step 200, both models solve 34 of the 99 tasks within four attempts under either scorer.
The agreement carries over to the test set, where each checkpoint generates 20 answers for each of 611 task/suite combinations and pass@1 is averaged equally over four suites.
With commercial Xcelium scoring, the OSS-trained model reaches 44.08\% and the Xcelium-trained model 44.24\%. For each model the scorer gap is below 0.2 percentage points.

Agreement in aggregate can still hide opposite verdicts on individual tasks.
For \texttt{always\_casez} (v1-human), Xcelium accepts all 20 commercial-trained answers, while Icarus rejects all 20 over \texttt{break} support. On \texttt{review2015\_fsmseq} (v1-machine), Icarus's acceptance of declarations after use reverses the direction of the gap.
VerilogEval uses Icarus~12, whose declaration handling differs from Icarus~13 in training.
Pinning the frontend version and language mode makes these syntax-dependent rewards reproducible.

\subsection{PPA Gains Across Synthesis Flows}
We train Qwen2.5-Coder-7B-Instruct for 200 CDPO updates with the ChipSeek correctness--PPA objective~\cite{chipseek} and the cap described in Section~\ref{sec:replay}.
One run uses Icarus and Yosys/OpenSTA, the other Xcelium and Genus. Each uses two H100 GPUs and 32 EDA slots. Adaptive sampling consumes 723 and 885 generation batches, respectively.
The runs share step-zero answers. Every 20 updates, each backend scores one answer per task on 200 NanGate45 tasks. From candidate and reference power ($P$), area ($A$), and delay ($D$) measured on that backend, we compute:
\[
  \mathrm{relative\ EDAP}
  = \frac{P}{P_{\mathrm{ref}}}
    \frac{A}{A_{\mathrm{ref}}}
    \frac{D}{D_{\mathrm{ref}}}.
\]

\begin{table}[!t]
  \centering
  \caption{Relative EDAP on 24 matched designs. Geometric means use each scoring backend's reference as $100\%$.}
  \label{tab:ppa-transfer}
  \footnotesize
  \setlength{\tabcolsep}{4pt}
  \renewcommand{\arraystretch}{1.12}
  \begin{tabular*}{\columnwidth}{@{\extracolsep{\fill}}lrrr@{}}
\toprule
 & \multicolumn{2}{c}{\textbf{Scoring backend}} & \\
\cmidrule(lr){2-3}
\textbf{Model} & \textbf{OSS (\%)} $\downarrow$ & \textbf{Comm. (\%)} $\downarrow$ & $\boldsymbol{\Delta}_{O-C}$ (p.p.) \\
\midrule
Initial & 106.04 & 101.56 & +4.48 \\
A (OSS-trained) & 106.11 & 101.19 & +4.93 \\
B (comm.-trained) & 104.34 & 100.10 & +4.24 \\
\midrule
$A-B$ (p.p.) & +1.78 & +1.09 & +0.69 \\
\bottomrule
\end{tabular*}

  \par\vspace{3pt}
  \begin{minipage}{\columnwidth}\footnotesize
  $\Delta_{O-C}$: OSS minus commercial; $A-B$: model difference.
  All differences use unrounded values, in percentage points (p.p.).
  \end{minipage}
\end{table}

Table~\ref{tab:ppa-transfer} compares the initial model and both step-200 checkpoints on 24 designs that pass simulation and have positive EDAP in every model/scorer combination, after excluding two known overflow errors.
OSS training yields a modest improvement under Genus (0.38 percentage points) even though its OSS-scored endpoint is nearly flat ($+0.07$\,p.p.).
Commercial training improves on the initial model under both scorers, by 1.46\,p.p. under Genus and 1.70\,p.p. under Yosys/OpenSTA.
The commercial-scored means move only slightly because Genus reports identical EDAP for all three model outputs on 20 of the 24 designs. Four designs thus account for the commercial-scored gains.

\paragraph{Which RTL gains survive synthesis?}
We compare answers from all checkpoints, including step zero, against each backend's synthesis of the reference RTL.
Across 33 designs, 178 jointly passing and measurable answers from the OSS run beat their reference under OSS synthesis. Genus preserves 34 of those gains, maps 98 answers to within $\pm0.1\%$ of reference EDAP, and scores 46 worse than the reference.

On the 16-by-8-bit unsigned divider (\texttt{vfc\_7551}), both step-200 models replace an unrolled shift-and-subtract loop with Verilog division and remainder. The OSS-trained answer passes the nonzero-divisor tests and has relative EDAP 0.93 under Yosys/OpenSTA and 0.30 under Genus. The initial answer fails simulation, so this gain falls outside Table~\ref{tab:ppa-transfer}.

The 8-bit multiplier (\texttt{vfc\_7527}, step~40) illustrates a reversal. Replacing shift-and-add with \texttt{A * B} lowers relative EDAP to 0.26 in the OSS flow. Under Genus, power rises by 8.5\%, outweighing small area and delay reductions and bringing relative EDAP to 1.02.

The most frequent outcome is a tie. In two such cases, Genus maps the reference and the rewrite to the same cells.

\noindent\begin{minipage}{\columnwidth}
In \texttt{vfc\_7623} (step 100), explicit pairwise equality replaces a vector expression:
\begin{lstlisting}[style=rtlcase]
// Reference: 25 pairwise comparisons
assign out = ~{{5{a}}, {5{b}}, {5{c}},
               {5{d}}, {5{e}}}
             ^ {5{a,b,c,d,e}};
// OSS-trained answer (two of 25 assignments)
out[23] = a == b;
out[19] = b == a;
\end{lstlisting}
\end{minipage}\par
The configured Yosys flow maps reference/answer to 20/10 XNOR cells (relative EDAP 0.25); Genus maps both to the same 10-cell circuit (1.00).
Yosys's \texttt{synth -flatten} variant also produces 10/10 cells, tying this difference to the configured optimization passes.

\noindent\begin{minipage}{\columnwidth}
In \texttt{vfc\_8053} (step 20), majority carry replaces the reference's two-half-adder construction:
\begin{lstlisting}[style=rtlcase]
// Reference: intermediate sum and carries
assign s1 = a ^ b;
assign c1 = a & b;
assign c2 = s1 & cin;
assign cout = c1 | c2;
// OSS-trained answer: majority carry
assign cout = (a & b) | (a & cin) | (b & cin);
\end{lstlisting}
\end{minipage}\par
Yosys maps both descriptions to five cells. The majority network uses two NAND gates where the reference uses a NAND and an inverter. Its 12.5\% shorter delay and lower power outweigh a 4.8\% area increase, giving relative EDAP 0.87 even with flattening enabled.
Genus uses one \texttt{FA\_X1} full adder in each netlist. Only the order of its symmetric inputs changes, leaving relative EDAP at 1.00.

These PPA results cover one training run per chain and modules with a median of 16 nonblank reference lines. Cross-scoring identifies the effect of changing the synthesizer on a fixed RTL answer. Estimating how the backend changes the policy would also require independent training seeds. On larger IP blocks the same paired test could measure whether the PPA advantage of a module-level rewrite survives integration.

\section{Conclusion}
\gym{} provides the first versatile, cross-task, and cross-toolchain EDA RL environment that replaces ad-hoc EDA evaluation scripts with reusable infrastructure for hardware design and verification.
Rather than rebuilding workspace handlers, resource guards, and reward scripts for every workflow, researchers can use a single candidate–feedback protocol across high-throughput parallel rollouts, multi-round repository repairs, and full-hierarchy SoC flows.
Streaming dispatch overlaps tool checks with rollout generation, while explicit memory budgets and formal/contract gates protect RL rewards from resource interference and invalid PPA shortcuts.

Beyond unifying single-chain workflows, \gym{}'s interchangeable backends bridge open-source training and commercial sign-off.
Paired cross-scoring shows that open-source simulation training preserves functional accuracy under Xcelium, while netlist comparison pinpoints which open-source PPA rewrites survive Genus and which are absorbed by commercial synthesis.
Teams with limited commercial licenses can thus train at scale on open tools and swap backends on saved revisions to audit gains before system integration.

\ifanonymous
  \clearpage
  \IEEEtriggeratref{14}
\else
  \IEEEtriggeratref{12}
\fi
\bibliographystyle{IEEEtran}
\bibliography{references}

\end{document}